\documentclass[11pt]{article}

\usepackage{relsize} 
\usepackage{acl2016}
\usepackage[colorlinks=true,linkcolor=black,citecolor=black,filecolor=black,urlcolor=black]{hyperref}

\usepackage{microtype}

\usepackage{titlesec}
\titleformat*{\subparagraph}{\itshape}
\titlespacing{\subparagraph}{%
  1em}{
  0pt}{
  1em}

\usepackage{lingmacros}
\newcommand{\exref}[1]{(\ref{#1})} 

\usepackage[shortlabels]{enumitem} 
\setlist{nolistsep}

\usepackage{natbib}

\usepackage[boxed]{algorithm2e} 

\usepackage[small,bf,skip=5pt]{caption}
\usepackage{sidecap} 
\usepackage{rotating}	

\usepackage{xspace}
\usepackage{xparse} 

\usepackage{textcomp}

\usepackage{framed}

\usepackage{listings}

\usepackage{amssymb}	
\usepackage{amsmath}

\usepackage{mathptmx}	
\usepackage[scaled=.8]{beramono}
\usepackage[scaled=.85]{helvet}
\usepackage[T1]{fontenc}
\usepackage[utf8x]{inputenc}

\usepackage{MnSymbol}	

\usepackage{latexsym}

\makeatletter
\DeclareTextCommandDefault{\fourdots}{%
.\kern\fontdimen3\font
.\kern\fontdimen3\font
.\kern\fontdimen3\font
.\kern\fontdimen3\font}
\makeatother

\usepackage{array}
\usepackage{longtable}
\usepackage{multirow}
\usepackage{booktabs} 
\usepackage{multicol}
\usepackage{footnote} 
\newcolumntype{H}{>{\setbox0=\hbox\bgroup}c<{\egroup}@{}} 

\usepackage{url}
\usepackage[usenames,dvipsnames,svgnames,table]{xcolor}

\usepackage[normalem]{ulem} 
\usepackage{colortbl}
\usepackage{graphicx}
\usepackage{subcaption}
\usepackage{tikz}
\usetikzlibrary{arrows,positioning,calc}

\usepackage{nameref}
\usepackage{cleveref}

\crefformat{part}{\S#2#1#3}
\crefformat{chapter}{\S#2#1#3}
\crefformat{section}{\S#2#1#3}
\crefformat{subsection}{\S#2#1#3}
\crefformat{subsubsection}{\S#2#1#3}
\crefformat{paragraph}{\P#2#1#3}
\crefformat{subparagraph}{\P#2#1#3}
\crefmultiformat{section}{\S#2#1#3}{ and~\S#2#1#3}{, \S#2#1#3}{, and~\S#2#1#3}
\crefmultiformat{subsection}{\S#2#1#3}{ and~\S#2#1#3}{, \S#2#1#3}{, and~\S#2#1#3}
\crefmultiformat{subsubsection}{\S#2#1#3}{ and~\S#2#1#3}{, \S#2#1#3}{, and~\S#2#1#3}
\crefmultiformat{paragraph}{\P\P#2#1#3}{ and~#2#1#3}{, #2#1#3}{, and~#2#1#3}
\crefmultiformat{subparagraph}{\P\P#2#1#3}{ and~#2#1#3}{, #2#1#3}{, and~#2#1#3}
\crefrangeformat{section}{\mbox{\S\S#3#1#4--#5#2#6}}
\crefrangeformat{subsection}{\mbox{\S\S#3#1#4--#5#2#6}}
\crefrangeformat{subsubsection}{\mbox{\S\S#3#1#4--#5#2#6}}
\crefrangeformat{paragraph}{\mbox{\P\P#3#1#4--#5#2#6}}
\crefrangeformat{subparagraph}{\mbox{\P\P#3#1#4--#5#2#6}}
\crefname{part}{Part}{Parts}
\Crefname{part}{Part}{Parts}
\crefname{chapter}{ch.}{ch.}
\Crefname{chapter}{Ch.}{Ch.}
\crefname{figure}{figure}{figures}
\crefname{subfigure}{figure}{figures}
\crefname{appsec}{appendix}{appendices}
\Crefname{appsec}{Appendix}{Appendices}
\crefname{algocf}{algorithm}{algorithms}
\Crefname{algocf}{Algorithm}{Algorithms}
\crefname{enums}{example}{examples}
\Crefname{enums}{Example}{Examples}
\crefname{enumsi}{example}{examples}
\Crefname{enumsi}{Example}{Examples}
\crefname{}{example}{examples} 
\Crefname{}{Example}{Examples}
\crefformat{enums}{(#2#1#3)}
\crefformat{enumsi}{(#2#1#3)}
\crefformat{}{(#2#1#3)}
\crefrangeformat{enums}{\mbox{(#3#1#4--#5#2#6)}}
\crefrangeformat{enumsi}{\mbox{(#3#1#4--#5#2#6)}}
\crefmultiformat{enumsi}{(#2#1#3}{, #2#1#3)}{, #2#1#3}{, #2#1#3)}
\crefrangemultiformat{enumsi}{(#3#1#4--#5#2#6}{, #3#1#4--#5#2#6)}{, #3#1#4--#5#2#6}{, #3#1#4--#5#2#6)}

\ifx\creflastconjunction\undefined%
\newcommand{\creflastconjunction}{, and\nobreakspace} 
\else%
\renewcommand{\creflastconjunction}{, and\nobreakspace} 
\fi%

\newcommand*{\Fullref}[1]{\hyperref[{#1}]{\Cref*{#1}: \nameref*{#1}}}
\newcommand*{\fullref}[1]{\hyperref[{#1}]{\cref*{#1}: \nameref{#1}}}
\newcommand{\fnref}[1]{footnote~\ref{#1}} 

\newcommand{\backtick}[0]{\textasciigrave}

\NewDocumentEnvironment{itmize}{}{\begin{itemize}[noitemsep]}{\end{itemize}}
\NewDocumentEnvironment{enumrate}{}{\begin{enumerate}[noitemsep]}{\end{enumerate}}
\let\Item\item
\renewcommand\enddescription{\endlist\global\let\item\Item}
\NewDocumentEnvironment{describe}{}{\renewcommand\item[1][]{\Item \textbf{##1:} }\begin{itemize}}{\end{itemize}}
\NewDocumentEnvironment{edescribe}{}{\renewcommand\item[1][]{\Item \textbf{##1:} }\begin{enumerate}}{\end{enumerate}}

\usepackage{venndiagram}

\usepackage{color}
\usepackage{bm}
\definecolor{orange}{rgb}{1,0.5,0}
\definecolor{mdgreen}{rgb}{0,0.6,0}
\definecolor{mdblue}{rgb}{0,0,0.7}
\definecolor{dkblue}{rgb}{0,0,0.5}
\definecolor{dkgray}{rgb}{0.3,0.3,0.3}
\definecolor{slate}{rgb}{0.25,0.25,0.4}
\definecolor{gray}{rgb}{0.5,0.5,0.5}
\definecolor{ltgray}{rgb}{0.7,0.7,0.7}
\definecolor{purple}{rgb}{0.7,0,1.0}
\definecolor{lavender}{rgb}{0.65,0.55,1.0}

\newcommand{\term}[1]{\textbf{#1}} 

\newcommand{\citeposs}[2][]{\citeauthor{#2}'s (\citeyear[#1]{#2})}
\newcommand{\Citeposs}[2][]{\Citeauthor{#2}'s (\citeyear[#1]{#2})}

\makeatletter
\renewcommand{\paragraph}{%
  \@startsection{paragraph}{4}%
  {\z@}{.2ex \@plus 1ex \@minus .2ex}{-.7em}%
  {\bfseries}%
}
\makeatother

\newcommand{\w}[1]{\textit{#1}}	
\newcommand{\p}[1]{\textbf{\textsf{#1}}} 
\newcommand{\lbl}[1]{\textsc{#1}} 
\newcommand{\sst}[1]{\lbl{#1}} 
\newcommand{\psst}[1]{\textcolor{mdgreen}{\sst{#1}}} 
\newcommand{\olbl}[1]{\textcolor{purple}{\textrm{#1}}} 

\newcommand{\dataset}[1]{\mbox{\textsc{#1}}}	

\newcommand{\tat}[0]{\textasciitilde}

\newcommand{\anonversion}[1]{}
\newcommand{\nonanonversion}[1]{#1}

\newcommand{\longversion}[1]{} 
\newcommand{\ignore}[1]{}

\title{A Corpus of Preposition Supersenses in English Web Reviews}
\date{}

\nonanonversion{\author{Nathan Schneider \\
  University of Edinburgh \\
  {\tt nschneid@inf.ed.ac.uk} \\\And
  Jena D. Hwang \\
  IHMC \\
  {\tt jhwang@ihmc.us}\\\And
  Vivek Srikumar \\
  University of Utah \\
  {\tt svivek@cs.utah.edu} \\\AND
  Meredith Green \quad Kathryn Conger \quad Tim O'Gorman \quad Martha Palmer\\
  University of Colorado at Boulder \\
  {\tt \{laura.green,kathryn.conger,timothy.ogorman,martha.palmer\}@colorado.edu}
\\}}

\aclfinalcopy 

\begin{document}
\maketitle
\begin{abstract}
We present the first corpus annotated with \textbf{preposition supersenses}, 
unlexicalized categories for semantic functions that can be marked by English prepositions \citep{schneider-15-pssts}.
That scheme improves upon its predecessors to better facilitate 
comprehensive manual annotation.
Moreover, unlike the previous schemes, the preposition supersenses are organized hierarchically.
Our data will be publicly released on the web upon publication.
\end{abstract}

\section{Introduction}

English prepositions exhibit stunning frequency and wicked polysemy. 
In the 450M-word COCA corpus \citep{coca}, 
11~prepositions are more frequent than the most frequent noun.\footnote{\url{http://www.wordfrequency.info/free.asp?s=y}}
In the corpus presented in this paper, prepositions account for 8.5\% of tokens (the top 11 prepositions comprise >6\% of all tokens). 
Far from being vacuous grammatical formalities, prepositions serve as essential linkers of meaning, 
and the few extremely frequent ones are exploited for many different functions (\cref{fig:polysemy}). 
For all their importance, however, prepositions have received 
relatively little attention in computational semantics, 
and the community has not yet arrived at a comprehensive and reliable
scheme for annotating the semantics prepositions in context (\cref{sec:bg}). 
We believe that such annotation of preposition functions is needed 
if preposition sense disambiguation systems are to be useful for 
downstream tasks---e.g., translation\footnote{This work focuses on English, but adposition and case systems 
vary considerably between languages, challenging second language learners 
and machine translation systems \citep{chodorow-07,shilon-12,hashemi-14}.} or semantic parsing \citep[cf.][]{dahlmeier-09,srikumar-11}.

\begin{figure}\small
\raggedright
\enumsentence{\label{ex:wildwood} I have been going \p{to}/\psst{Destination} the Wildwood\_,\_NJ \p{for}/\psst{Duration} 
over 30 years \p{for}/\psst{Purpose} summer\tat vacations\\[-.5ex]}
\enumsentence{It is close \p{to}/\psst{Location} bus\_lines \p{for}/\psst{Destination} Opera\_Plaza\\[-.5ex]}
\enumsentence{I was looking\tat \p{to}/\olbl{\backtick i} bring a customer \p{to}/\psst{Destination} 
their lot \p{to}/\psst{Purpose} buy a car}
\caption{Preposition supersense annotations illustrating polysemy of \p{to} and \p{for}. 
Note that both can mark a \psst{Destination} or \psst{Purpose}, while there are other functions that do not overlap.
The syntactic complement use of infinitival \p{to} is tagged as \olbl{\backtick i}. The \p{over} token in \exref{ex:wildwood} receives the label \psst{Approximator}. See \cref{sec:annoscheme} for details.}
\label{fig:polysemy}
\end{figure}

This paper describes a new corpus fully annotated with preposition supersenses 
(hierarchically organized unlexicalized classes). 
We note that none of the existing English corpora annotated with 
preposition semantics, on which existing disambiguation models have been trained and evaluated, 
are both \emph{comprehensive} (describing all preposition types and tokens) 
and \emph{double-annotated} (to attenuate subjectivity in the annotation scheme and measure inter-annotator agreement). 
As an alternative to fine-grained sense annotation for individual prepositions---which is difficult and limited by 
the coverage and quality of a lexicon---%
we instead train human annotators to label \textbf{preposition supersenses}, 
reporting the first inter-annotator agreement figures for this task.
We comprehensively annotate English preposition tokens in a \textbf{corpus} 
of web reviews and examine the distribution of their supersenses, and improve upon the supersense hierarchy as necessitated by the data encountered during the annotation process. 
Our annotated corpus will be publicly released at the time of publication.

\section{Background and Motivation}\label{sec:bg}

Theoretical linguists have puzzled over questions such as how individual prepositions 
can acquire such a broad range of meanings---and to what extent those meanings are 
systematically related \citep[e.g.,][]{brugman-81,lakoff-87,tyler-03,odowd-98,saint-dizier-06,lindstromberg-10}. 

Prepositional polysemy has also been recognized 
as a challenge for AI \citep{herskovits-86} and natural language processing, 
motivating semantic disambiguation systems \citep{ohara-03,ye-07,hovy-10,srikumar-13}.
Training and evaluating these requires semantically annotated corpus data. 
Below, we comment briefly on existing resources and why (in our view) 
a new resource is needed to ``road-test'' 
an alternative, hopefully more scalable, 
semantic representation for prepositions.

\subsection{Existing Preposition Corpora}

Beginning with the seminal resources from The Preposition Project \citep[TPP;][]{litkowski-05}, the computational study of preposition semantics has been fundamentally grounded in corpus-based lexicography centered around individual preposition types.
Most previous datasets of preposition semantics at the token level \citep{litkowski-05,litkowski-07,dahlmeier-09,tratz-09,srikumar-13-inventory} only cover high-frequency prepositions (the 34 represented in the SemEval-2007 shared task based on TPP, or a subset thereof).\footnote{A further limitation of the SemEval-2007 dataset is the way in which it was sampled: illustrative tokens from a corpus were manually selected by a lexicographer. As \citep{litkowski-14} showed, a disambiguation system trained on this dataset will therefore be biased and perform poorly on an ecologically valid sample of tokens.}

We sought a scheme that would facilitate \emph{comprehensive} semantic annotation 
of all preposition tokens in a corpus: 
thus, it would have to cover the full range of usages possible for the full range of English preposition types.
The recent TPP PDEP corpus \citep{litkowski-14,litkowski-15} comes closer to this goal, as it consists of randomly sampled tokens for over 300 types. 
However, sentences were sampled separately for each preposition, 
so there is only one annotated preposition token per sentence. 
By contrast, we will fully annotate documents for all preposition tokens.
No inter-annotator agreement figures have been reported for the PDEP data to indicate its quality, 
or the overall difficulty of token annotation with TPP senses across a broad range of prepositions.

\begin{figure*}
\includegraphics[width=\textwidth]{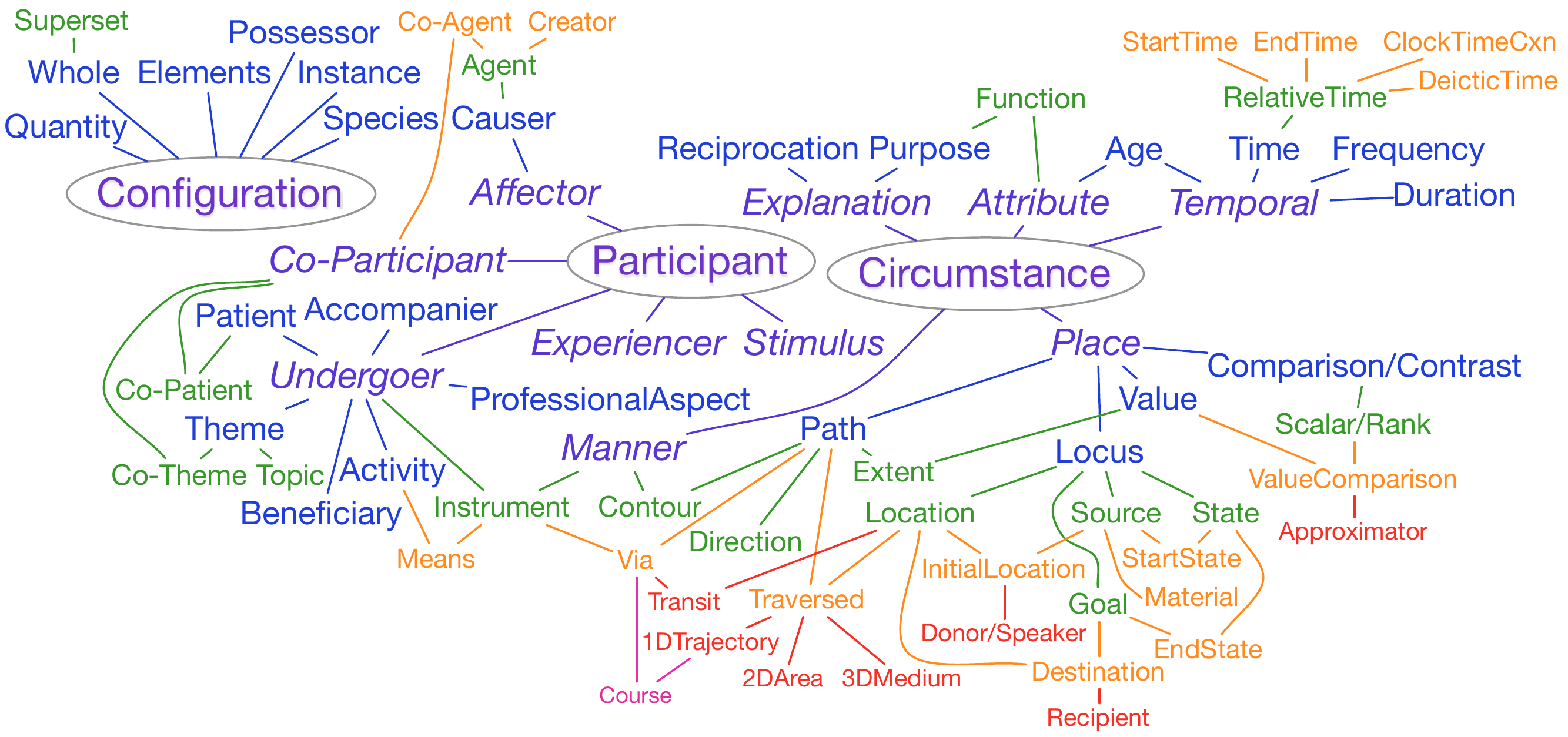}
\caption{Supersense hierarchy used in this work \citep[adapted from][]{schneider-15-pssts}. Circled nodes are roots (the most abstract categories); 
subcategories are shown above and below. Each node's color and formatting reflect its depth.}
\label{fig:hierarchy}
\end{figure*}

\subsection{Supersenses}

From the literature on other kinds of supersenses, there is reason to believe
that token annotation with \term{preposition supersenses} \citep{schneider-15-pssts} 
will be more scalable and useful than senses. 
The term \term{supersense} has been applied to lexical semantic classes that label 
a large number of word types (i.e., they are unlexicalized). 
The best-known supersense scheme draws on two inventories---one for nouns and one for verbs---which 
originated as a high-level partitioning of senses in WordNet \citep{miller-90}.
A scheme for adjectives has been proposed as well \citep{tsvetkov-14}.

One argument advanced in favor of supersenses is that they provide a coarse level of generalization 
for essential contextual distinctions---such as artifact vs.~person for \emph{chair}, 
or temporal vs.~locative \p{in}---without being so fine-grained that systems cannot learn them \citep{ciaramita-06}. 
A similar argument applies for \emph{human} learning as pertains to rapid, cost-effective, and open-vocabulary 
annotation of corpora: an inventory of dozens of categories (with mnemonic names) 
can be learned and applied to unlimited vocabulary without having to refer to dictionary definitions \citep{schneider-12}. 
Like with WordNet for nouns and verbs, the same argument holds for prepositions: 
TPP-style sense annotation requires familiarity 
with a different set of (often highly nuanced) distinctions for each preposition type. 
For example, \p{in} has 15 different TPP senses, among them 
\p{in 10(7a)} `indicating the key in which a piece of music is written: \emph{Mozart's Piano Concerto in E flat}'.

Supersenses have been exploited for a variety of tasks \citep[e.g.,][]{agirre-08,tsvetkov-13,tsvetkov-15},
and full-sentence noun and verb taggers have been built for several languages \citep{segond-97,johannsen-14,picca-08,martinez-15,schneider-13,dimsum-16}. 
They are typically implemented as sequence taggers. In the present work, we extend a corpus 
that has already been hand-annotated with noun and verb supersenses, 
thus raising the possibility of systems that can learn all three kinds of supersenses jointly \citep[cf.][]{srikumar-13}.

\subsection{PrepWiki}

\Citeposs{schneider-15-pssts} preposition supersense scheme is described in detail in a lexical resource, PrepWiki,\nonanonversion{\footnote{\url{http://tiny.cc/prepwiki}}} 
which records associations between supersenses and preposition types. 
Hereafter, we adopt the term \term{usage} for a pairing of a preposition type 
and a supersense label---e.g., \p{at}/\psst{Time}.
Usages are organized in PrepWiki via (lexicalized) \term{senses} from the TPP lexicon.
The mapping is many-to-many, as senses and supersenses capture different generalizations.
(TPP senses, being lexicalized, are more numerous and generally finer-grained, but in some cases lump together 
functions that receive different supersenses, as in the sense \p{for 2(2)} `affecting, with regard to, or in respect of'.)
Thus, for a given preposition, a sense may be mapped to multiple usages, 
and vice versa.

\subsection{The Supersense Hierarchy}

Of the four supersense schemes mentioned above, 
\citeposs{schneider-15-pssts} inventory for prepositions 
(which improved upon the inventory of \cite{srikumar-13-inventory}) is unique in being hierarchical. 
It is an inheritance hierarchy (see \cref{fig:hierarchy}): characteristics of higher-level categories are asserted to apply to their descendants.
Multiple inheritance is used for cases of overlap: e.g., \psst{Destination} 
inherits from both \psst{Location} (because a destination is a point in physical space) 
and \psst{Goal} (it is the endpoint of a concrete or abstract path).

The structure of the hierarchy was modeled after VerbNet's hierarchy of thematic roles \citep{bonial-11,hwang-14}.
But there are many additional categories: some are refinements of the VerbNet roles (e.g., subclasses of \psst{Time}),
while others have no VerbNet counterpart because they do not pertain to core roles of verbs.
The \psst{Configuration} subhierarchy, which is used for \p{of} and other prepositions when they relate two nominals, 
is a good example.

\section{Corpus Annotation}\label{sec:corpus}

\subsection{Annotating Preposition Supersenses}\label{sec:annoscheme}

\paragraph{Source data.} \label{sec:source_data}
We fully annotated the \dataset{Reviews} section of the English Web Treebank \citep{ewtb}, 
selected because it had previously been annotated for multiword expressions and 
noun and verb supersenses \citep{schneider-14-mwecorpus,schneider-15-nvsst}. 
The corpus consists of 55,579 tokens organized into 3812~sentences and 
723~documents, with gold tokenization and PTB-style POS tags.

\paragraph{Identifying preposition tokens.}
TPP, and therefore PrepWiki, contains senses for canonical prepositions, i.e., those used transitively 
in the [$_\text{PP}$~P~NP] construction.
Taking inspiration from \citet{pullum-02}, PrepWiki further assigns supersenses to 
spatiotemporal particle uses of \p{out}, \p{up}, \p{away}, \p{together}, etc., 
and subordinating uses of \p{as}, \p{after}, \p{in}, \p{with}, etc.\ 
(including infinitival \p{to} and infinitival-subject \p{for}, as in 
\w{It took over 1.5 hours \p{for} our food \p{to} come out}).\footnote{PrepWiki 
does not include subordinators\slash complementizers that cannot take NP complements: \emph{that}, \emph{because}, \emph{while}, \emph{if}, etc.}

\subparagraph{Non-supersense labels.}\label{sec:nonsupersense} These are used  
where the heuristics fail (sometimes due to a POS tagging error)
or where the preposition serves a special syntactic function 
not captured by the supersense inventory. The most frequent is \olbl{\backtick i}, 
which applies only to infinitival \p{to} tokens that are not \psst{Purpose} or \psst{Function} adjuncts.\footnote{See \cref{fig:polysemy} 
for examples from the corpus. \w{I want/love/try \p{to} eat cookies} and \w{\p{To} love is \p{to} suffer} would qualify as \olbl{\backtick i}; 
\w{a shoulder \p{to} cry on} would qualify as \psst{Function}.}
The label \olbl{\backtick d} applies to discourse expressions; the
unqualified backtick (\olbl{\backtick}) applies to miscellaneous cases 
such as infinitival-subject \p{for} 
and both prepositions in the \p{as}-\p{as} comparative construction 
(\w{\p{as} wet \p{as} water}; \w{\p{as} much cake \p{as} you want}).

\subparagraph{Multiword expressions.}
\Cref{fig:mwes} shows how prepositions can interact with multiword expressions (MWEs).
An MWE may function holistically as a preposition: PrepWiki treats these as multiword prepositions.
An idiomatic phrase may be headed by a preposition, in which case we assign it a preposition supersense 
or tag it as a discourse expression (\olbl{\backtick d}). 
Finally, a preposition may be embedded within an MWE (but not its head): we do not use a preposition supersense 
in this case, though the MWE as a whole may already be tagged with a verb supersense.

\subparagraph{Heuristics.}
The annotation tool uses heuristics to detect candidate preposition tokens in each sentence 
given its POS tagging and MWE annotation.
A \emph{single-word expression} is included if:
\begin{itemize}
\item it is tagged as a verb particle (\lbl{rp}) or infinitival \p{to} (\lbl{to}), or
\item it is tagged as a transitive preposition or subordinator (\lbl{in}) or adverb (\lbl{rb}), and the word is listed in PrepWiki (or the spelling variants list).
\end{itemize}
A strong \emph{MWE} instance is included if:
\begin{itemize}
\item the MWE begins with a word that matches the single-word criteria (idiomatic PP), or
\item the MWE is listed in PrepWiki (multiword preposition). 
\end{itemize}

\begin{figure}\small
\raggedright
\enumsentence{\label{ex:because-of}\fbox{\p{Because\_of}/\psst{Explanation}} the ants 
I dropped them \p{to}/\psst{EndState} a 3\_star .\\[-.5ex]}
\enumsentence{\label{ex:to-go}I was told \p{to}/\olbl{\backtick i} take my coffee 
\fbox{\p{to}\_go/\psst{Manner}} if I wanted \p{to}/\olbl{\backtick i} finish it .\\[-.5ex]}
\enumsentence{\label{ex:to-boot}\p{With}/\psst{Attribute} higher \p{than}/\psst{Scalar/Rank} 
average prices \fbox{\p{to}\_boot/\olbl{\backtick d}} !\\[.5ex]}
\enumsentence{\label{ex:in-mwe}I worked\tat \p{with}/\psst{ProfessionalAspect} Sam\_Mones 
who \fbox{took\_ great \_care\_\p{of}} me .}
\caption{Prepositions involved in multiword expressions. 
\cref{ex:because-of}~Multiword preposition \p{because of} (others include \p{in~front~of}, \p{due~to}, \p{apart~from}, and \p{other~than}). 
\cref{ex:to-go}~PP idiom: the preposition supersense applies to the MWE as a whole.
\cref{ex:to-boot}~Discourse PP idiom: instead of a supersense, expressions serving a discourse function are tagged as \olbl{\backtick d}.
\cref{ex:in-mwe}~Preposition within a multiword expression: the expression is headed by a verb, so it receives 
a verb supersense (not shown) rather than a preposition supersense.}
\label{fig:mwes}
\end{figure}

\begin{figure*}[t]
\includegraphics[width=\textwidth,trim=0 275 0 0,clip]{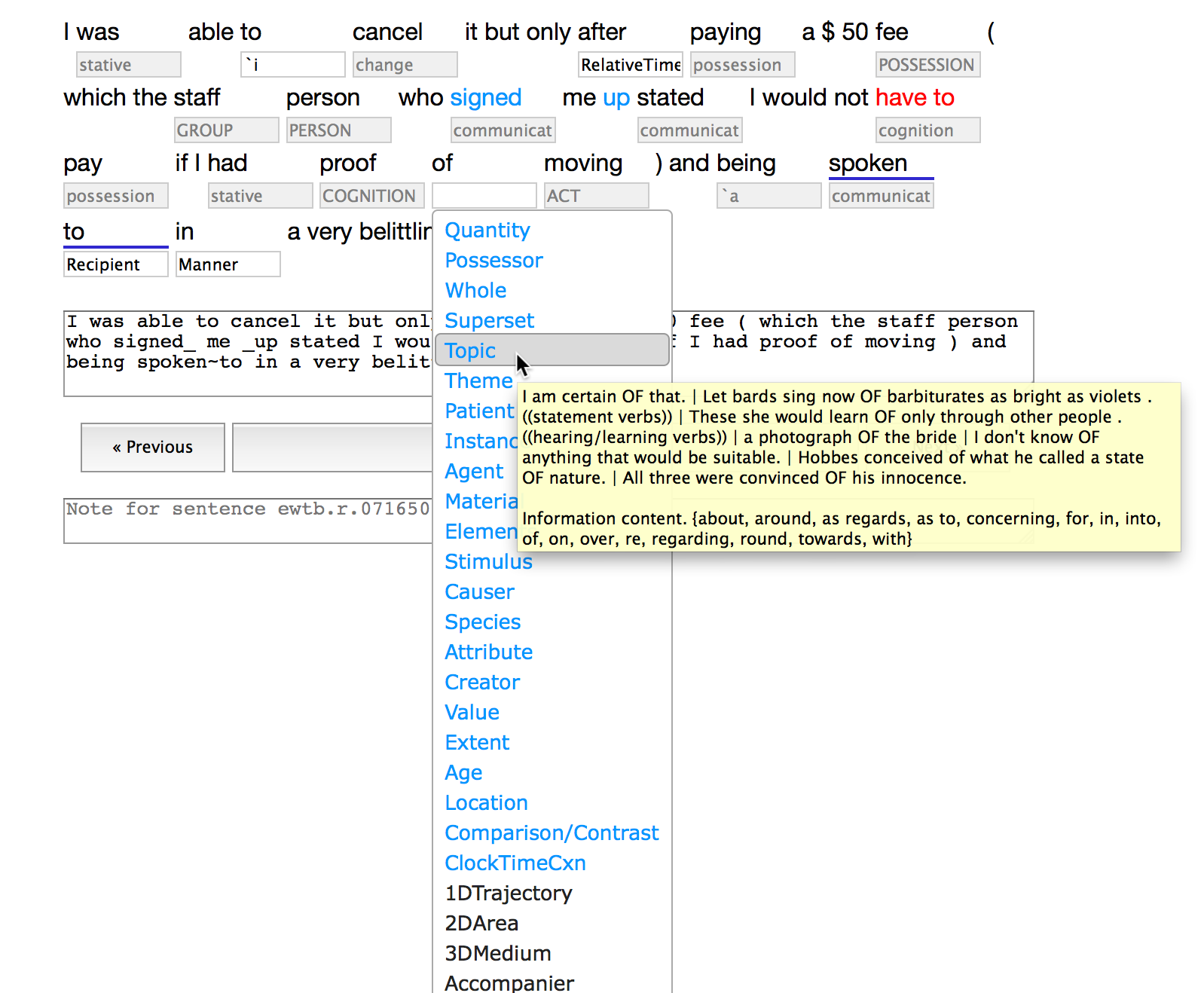}
\caption{Supersense annotation interface, developed in-house. 
The main thing to note is that preposition, noun, and verb supersenses are stored in text boxes below 
the sentence. A dropdown menu displays the full list of preposition supersenses, 
starting with those with PrepWiki mappings to the preposition in question. 
Hovering the mouse over a menu item displays a tooltip with PrepWiki examples 
of the usage (if applicable) and a general definition of the supersense.}
\label{fig:interface}
\end{figure*}

\paragraph{Annotation task.}
Annotators proceeded sentence by sentence, working in a custom web interface (\cref{fig:interface}).
For each token matched by the above heuristics, 
annotators filled in a text box with the contextually appropriate label. 
A dropdown menu showed the list of preposition supersenses and non-supersense labels, 
starting with labels known to be associated with the preposition being annotated. 
Hovering over a menu item would show example sentences to illustrate the usage in question, 
as well as a brief definition of the supersense.
This preposition-specific rendering of the dropdown menu---supported by data 
from PrepWiki---was 
crucial to reducing the overhead of annotation (and annotator training) 
by focusing the annotator's attention on the relevant categories/usages.
New examples were added to PrepWiki as annotators spotted coverage gaps.
The tool also showed the multiword expression annotation of the sentence, 
which could be modified if necessary to fit PrepWiki's conventions 
for multiword prepositions.

\subsection{Quality Control}

\paragraph{Annotators.} 
Annotators were selected from undergraduate and graduate linguistics students\nonanonversion{ at the University of 
Colorado at Boulder}.
All annotators had prior experience with semantic role labeling.
Every sentence was independently annotated by two annotators, and disagreements 
were subsequently adjudicated by a third, ``expert'' annotator. 
There were two expert annotators\nonanonversion{, both authors of this paper}.

\paragraph{Training.} 200~sentences were set aside for training annotators. 
Annotators were first shown how to use the preposition annotation tool 
and instructed on the supersense distinctions for this task. 
Annotators then completed a training set of 100 sentences. 
An adjudicator evaluated the annotator's annotations, providing feedback 
and assigning another 50--100 training instances if necessary.

Inter-annotator agreement (IAA) measures are useful in quantifying annotation ``reliability'', 
i.e., indicating how trustworthy and reproducible the process is (given guidelines, training, tools, etc.).
Specifically, IAA scores can be used as a diagnostic for the reliability of (i)~individual annotators 
(to identify those who need additional training/guidance); 
(ii)~the annotation scheme and guidelines (to identify problematic phenomena requiring further documentation 
or substantive changes to the scheme); 
(iii)~the final dataset (as an indicator of what could reasonably be expected of an automatic system).

\paragraph{Individual annotators.}  
The main annotation was divided into 34~batches of 100~sentences. 
Each batch took on the order of an hour for an annotator to complete.
We monitored original annotators' IAA throughout the annotation process 
as a diagnostic for when to intervene in giving further guidance.
Original IAA for most of these batches fell between 60\% and 78\%, depending on factors such as 
the identities of the annotators and when the annotation took place 
(annotator experience and PrepWiki documentation improved over time).\footnote{Specifically, the agreement rate among tokens 
where both annotators assigned a preposition supersense was between 82\% and 87\% for 4 batches; 
72\% and 78\% for 11; 60\% and 70\% for 17; and below 60\% for 2.
This measure did not award credit for agreement on non-supersense labels 
and ignored some cases of disagreement on the MWE analysis.\label{fn:iaameasure}}  
These rates show that it was not an easy annotation task, 
though many of the disagreements were over slight distinctions in the hierarchy (such as \psst{Purpose} vs.~\psst{Function}).

\paragraph{Guidelines.} Though \citet{schneider-15-pssts} conducted pilot annotation in constructing the supersense inventory, 
our annotators found a few details of the scheme to be confusing.
Informed by their difficulties and disagreements, we therefore made several minor 
improvements to the preposition supersense categories and hierarchy structure.
For example, the supersense categories for partitive constructions proved persistently problematic, 
so we adjusted their boundaries and names.
We also improved the high-level organization of the original hierarchy, 
clarified some supersense descriptions, 
and removed the miscellaneous \psst{Other} supersense.

\begin{SCfigure*}\small\centering
\includegraphics[width=.62\textwidth]{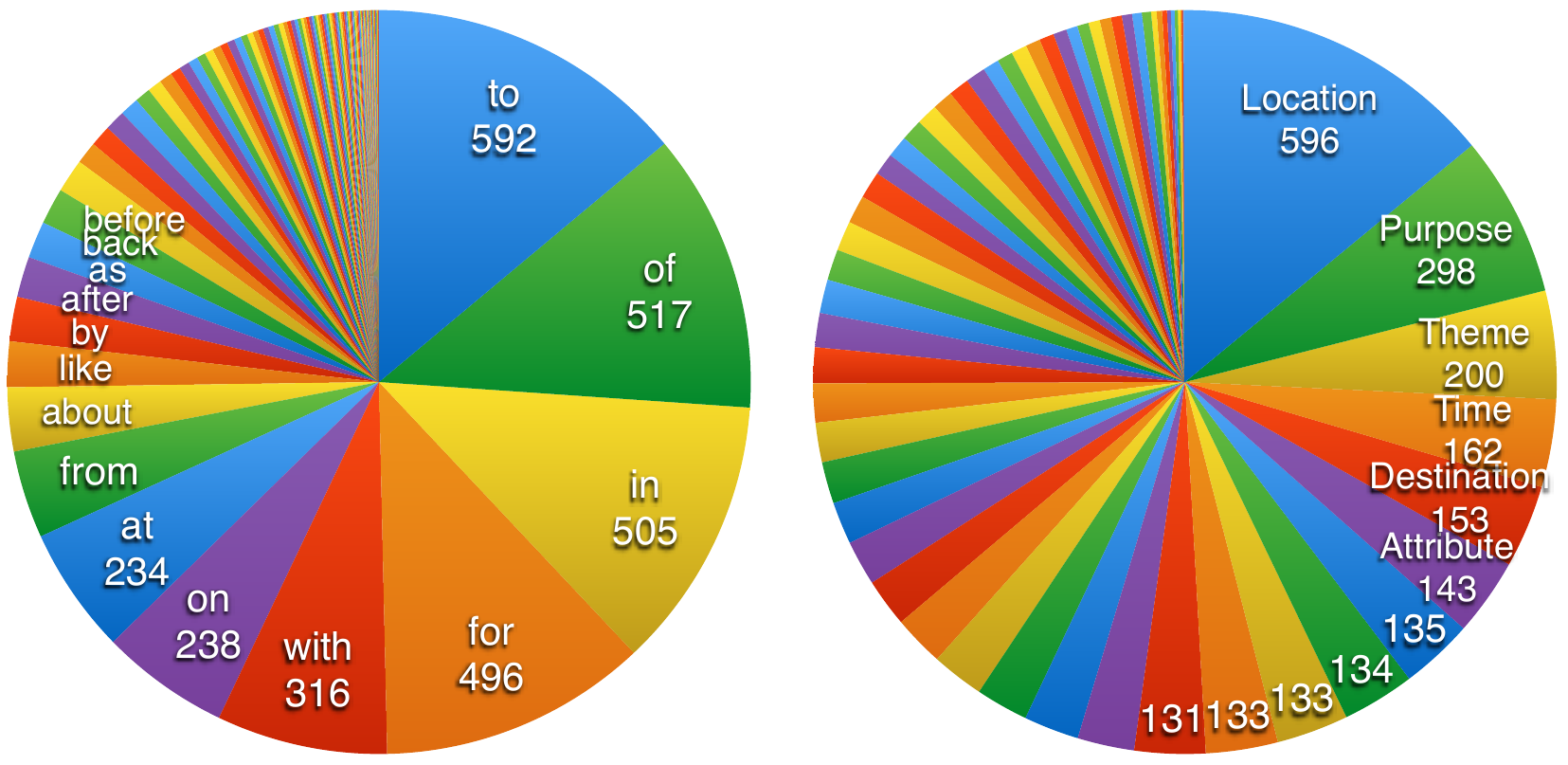}
\caption{Distributions of preposition types and supersenses for the 4,250 supersense-tagged 
preposition tokens in the corpus. In total, 114~prepositions and 63~supersenses are attested. 
Observe that just 9 prepositions account for 75\% of tokens, whereas the head of the supersense 
distribution is much smaller.}
\label{fig:pies}
\end{SCfigure*}

\paragraph{Revisions.} 
The changes to categories/guidelines noted in the previous paragraph 
required a small-scale post hoc revision to the annotations, which was performed by the expert annotators.
Some additional post hoc revisions were performed to improve consistency; 
e.g., some anomalous multiword expression annotations involving prepositions were fixed.\nonanonversion{\footnote{In particular, 
many of the borderline prepositional verbs were revised according to the guidlines outlined at 
\url{https://github.com/nschneid/nanni/wiki/Prepositional-Verb-Annotation-Guidelines}.}}

\paragraph{Adjudication reliability.} Because sentences were adjudicated by one of two expert annotators, 
we can estimate the dataset's adjudication reliability---roughly, the expected proportion of tokens 
that would have been labeled the same way if adjudicated by the other expert---by measuring IAA 
on a sample independently annotated 
by both experts.\footnote{These sentences were then jointly adjudicated by the experts to arrive at a final version.} 
Applying this procedure to 203~sentences annotated late in the process
(using the measure described in \fnref{fn:iaameasure})
gives an agreement rate of $276/313=88\%$.\footnote{For completeness, Cohen's $\kappa = .878$. 
It is almost as high as raw agreement because the expected agreement rate is very low---but keep in mind 
that $\kappa$'s model of chance agreement does not take into account preposition types or the 
fact that a relatively small subset of labels were suggested for most prepositions.
On the 4~most frequent prepositions in the sample, \emph{per-preposition} $\kappa$ is 
.84 for \p{for}, 1.0 for \p{to}, .59 for \p{of}, and .73 for \p{in}.}
It is difficult to put an exact quality figure on a dataset that was developed over a period of time 
and with the involvement of many individuals; however, 
the fact that the expert-to-expert adjudication estimate approaches 90\% 
despite the large number of labels suggests that the data can serve as a reliable resource for 
training and benchmarking disambiguation systems.

\subsection{Resulting Corpus}

4250 tokens have preposition supersenses. 
Their distribution appears in \cref{fig:pies}.
Over~75\% of tokens belong to the top 10 preposition types, while the supersense distribution is closer to uniform.
1170~tokens are labeled as \psst{Location}, \psst{Path}, or a subtype thereof: these can roughly be described as spatial. 
528~come from the \psst{Temporal} subtree of the hierarchy, and 452 from the \psst{Configuration} subtree.
Thus, fully half the tokens (2100) mark non-spatiotemporal participants and circumstances.

Of the 4250 tokens, 582 are MWEs (multiword prepositions and/or PP idioms).%
\footnote{\label{fn:ppidioms}For the purpose of counting prepositions by type, we split up supersense-tagged 
PP idioms such as those shown in \exref{ex:to-go} and \exref{ex:to-boot} by taking the longest prefix of words 
that has a PrepWiki entry to be the preposition.}
A further 588 have non-supersense labels: 
484~\olbl{\backtick i}, 83~\olbl{\backtick d}, and 21~\olbl{\backtick}.

\begin{figure*}[t]
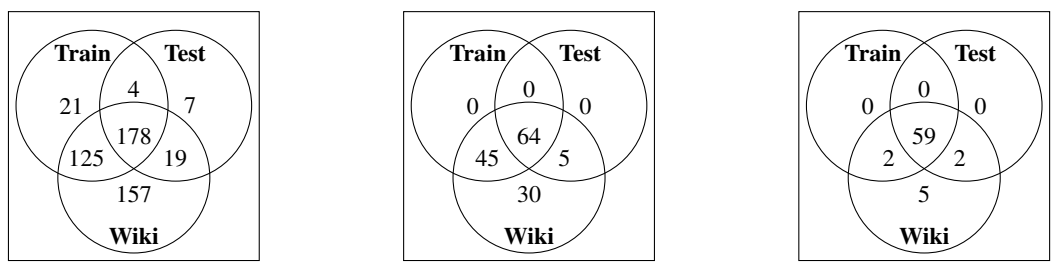
\small\centering
\begin{subfigure}[t]{.35\textwidth}\centering
\begin{venndiagram3sets}[labelOnlyA=21,labelOnlyB=7,labelOnlyC=157,
 labelOnlyAB=4,labelOnlyAC=125,labelOnlyBC=19,labelABC=178,
labelA=\textbf{Train\phantom{xl}\vphantom{\large l}},labelB=\textbf{\phantom{xl}\vphantom{\large l}Test},labelC=\textbf{Wiki},radius=1cm,overlap=.9cm,hgap=0.1cm,vgap=0.1cm]
\end{venndiagram3sets}
\caption{\textbf{Usages (preposition type + supersense).} 
E.g., \p{after}/\psst{Explanation} and \p{into}/\psst{EndState} are recorded in the wiki 
and attested in the training data but not the test data.
(Recall that PrepWiki was updated over the course of annotation, 
so these figures are not intended to predict its coverage of unseen data. We refrained from adding to PrepWiki 
a few usages that appeared infrequently in the data and seemed grammatically marginal or had a debatable supersense annotation.)}
\end{subfigure}
\hfill
\begin{subfigure}[t]{.26\textwidth}\centering
\begin{venndiagram3sets}[labelOnlyA=0,labelOnlyB=0,labelOnlyC=30,
 labelOnlyAB=0,labelOnlyAC=45,labelOnlyBC=5,labelABC=64,
labelA=\textbf{Train\phantom{xl}\vphantom{\large l}},labelB=\textbf{\phantom{xl}\vphantom{\large l}Test},labelC=\textbf{Wiki},radius=1cm,overlap=.9cm,hgap=0.1cm,vgap=0.1cm]
\end{venndiagram3sets}
\caption{\textbf{Prepositions with $\geq$1 usage.} Examples occurring in only one of the data splits 
include \p{despite}, \p{in spite of}, \p{onto}, \p{via}, and \p{on top of}. 
The 30~prepositions listed only for the wiki only counts
those with at least one mapped supersense.}
\end{subfigure}
\hfill
\begin{subfigure}[t]{.35\textwidth}\centering
\begin{venndiagram3sets}[labelOnlyA=0,labelOnlyB=0,labelOnlyC=5,
 labelOnlyAB=0,labelOnlyAC=2,labelOnlyBC=2,labelABC=59,
labelA=\textbf{Train\phantom{xl}\vphantom{\large l}},labelB=\textbf{\phantom{xl}\vphantom{\large l}Test},labelC=\textbf{Wiki},radius=1cm,overlap=.9cm,hgap=0.1cm,vgap=0.1cm]
\end{venndiagram3sets}
\caption{\textbf{Supersenses with $\geq$1 usage.} \psst{Creator}, \psst{Co-Patient}, \psst{Transit}, \psst{Temporal}, and \psst{3DMedium} 
are associated with usages in the wiki, but these usages are rare and did not appear in our data. 
7~supersenses---\psst{Configuration}, \psst{Participant}, \psst{Affector}, \psst{Undergoer}, \psst{Place}, \psst{Path}, and \psst{Traversed}---are abstractions 
intended solely for structuring the wiki; they are not used directly to label prepositions either in the wiki or in the data.}
\end{subfigure}
\caption{Venn diagrams counting \emph{types} of usages, prepositions, and supersenses in the data and wiki.}
\label{fig:venn}
\end{figure*}

\subsection{Splits}

To facilitate future experimentation on a standard benchmark, 
we partitioned our data into training and test sets. 
We randomly sampled 447 \emph{sentences} 
(4,073~total tokens and $N=950=19.6\%$ of preposition instances) 
for a held-out test set, leaving 3,888~preposition instances for training.\footnote{Excluding \olbl{\backtick i} and 
\olbl{\backtick other} instances, the supersense-labeled prepositions amount to 3,397 training  
and 853 test instances.} 
The sampling was stratified by preposition supersense so as to encourage a 
reasonable balance for the rare labels; e.g., supersenses that occur twice are split so that 
one instance is assigned to the training set and one to the test set.\footnote{The sampling algorithm 
considered supersenses in increasing order of frequency: for each supersense $\ell$ 
having $n_{\ell}$ instances, enough sentences were assigned to the test 
set to fill a minimum quota of $\lceil.195 n_{\ell}\rceil$ tokens for that supersense 
(and remaining unassigned sentences 
containing that supersense were placed in the training set). 
Relative to the training set, 
the test set is skewed slightly in favor of rarer supersenses. A small number of annotation errors 
were corrected subsequent to determining the splits. Entire sentences were sampled 
to facilitate future studies involving joint prediction over the full sentence.}
\Cref{fig:venn} shows, at a type level, the extent of overlap between the training set, test set, and PrepWiki.
61~preposition supersenses are attested in the training data, while 14~are unattested.

\section{Conclusion}

We have introduced a new lexical semantics corpus that
disambiguates prepositions with hierarchical supersenses.
Because it is comprehensively annotated over full documents, it offers insights 
into the semantic distribution of prepositions. 
The corpus will further facilitate the development of automatic preposition disambiguation systems.

\nonanonversion{\section*{Acknowledgments}

We thank our annotators---%
Evan Coles-Harris, Audrey Farber, Nicole Gordiyenko, Megan Hutto, 
Celeste Smitz, and Tim Watervoort---%
as well as Ken Litkowski, Michael Ellsworth, Orin Hargraves, and Susan Brown for helpful discussions.
This research was supported in part by a Google research grant for Q/A PropBank Annotation.}

\bibliographystyle{aclnat}
\begingroup
	\small
	\setlength{\bibsep}{0.2pt}
	\bibliography{psst}
\endgroup

\end{document}